\documentclass[letterpaper,twocolumn,fleqn]{article} 

\usepackage{ist}
\usepackage{algorithm2e}

\usepackage[table]{xcolor}

\usepackage{graphicx}
\usepackage{subcaption}
\usepackage{float}

\usepackage{url}

\title{Detecting GAN generated Fake Images using Co-occurrence Matrices}

\author{Lakshmanan Nataraj; Mayachitra Inc., Santa Barbara, California, USA\\
Tajuddin Manhar Mohammed; Mayachitra Inc., Santa Barbara, California, USA\\
Shivkumar Chandrasekaran;  Mayachitra Inc., Santa Barbara, California, USA\\
Arjuna Flenner; Naval Air Warfare Center Weapons Division, China Lake, California, USA\\
Jawadul H. Bappy; JD.com\\
Amit K. Roy-Chowdhury; University of California, Riverside, California, USA\\
B. S. Manjunath; Mayachitra Inc., Santa Barbara, California, USA}

\date{} 

\begin{document} 

\maketitle 

\thispagestyle{empty} 


\begin{abstract}

The advent of Generative Adversarial Networks (GANs) has brought about completely novel ways of transforming and manipulating pixels in digital images.
GAN based techniques such as Image-to-Image translations, DeepFakes, and other automated methods have become increasingly popular in creating fake images.  
In this paper, we propose a novel approach to detect GAN generated fake images using a combination of co-occurrence matrices and deep learning. We extract co-occurrence matrices on three color channels in the pixel domain and train a model using a deep convolutional neural network (CNN) framework. Experimental results on two diverse and challenging GAN datasets comprising more than 56,000 images based on unpaired image-to-image translations (cycleGAN~\cite{zhu2017unpaired}) and facial attributes/expressions (StarGAN~\cite{choi2018stargan}) show that our approach is promising and achieves more than 99\% classification accuracy in both datasets. Further, our approach also generalizes well and achieves good results when trained on one dataset and tested on the other.  
\end{abstract}


\section{Introduction}
\label{sec:intro}


Recent advances in Machine Learning and Artificial Intelligence have made it tremendously easy to create and synthesize digital manipulations in images and videos. 
In particular, Generative Adversarial Networks (GANs)~\cite{goodfellow2014generative} have been one of the most promising advancements in image enhancement and manipulation. 
Due to the success of using GANs for image editing, it is now possible to use a combination of GANs and off-the-shelf image-editing tools to modify digital images to such an extent that it has become difficult to distinguish doctored images from normal ones.
The field of digital Image Forensics develops tools and techniques to detect manipulations in digital images such as splicing, resampling and copy move, but the efficacy and robustness of these tools on GAN generated images is yet to be seen. 
To address this, we propose a novel method to automatically identify GAN generated fake images using techniques that have been inspired from classical steganalysis.

The seminal work on GANs\cite{goodfellow2014generative} cast the machine learning field of generative modeling as a game theory optimization problem.  
GANs contain two networks - the first network is a generative network that can generate fake images and the second network is a discirminative network that determines if an image is real or fake. 
Encoded in the GAN loss function is a min-max game which creates a competition between the generative and discriminative networks.  
As the discriminative network becomes better at distinguishing between real and fake images, the generative model becomes better at generating fake images.

\begin{figure*}[!ht]
  \centering
  \begin{subfigure}{\textwidth}
  \centering
  \includegraphics[scale=0.68]{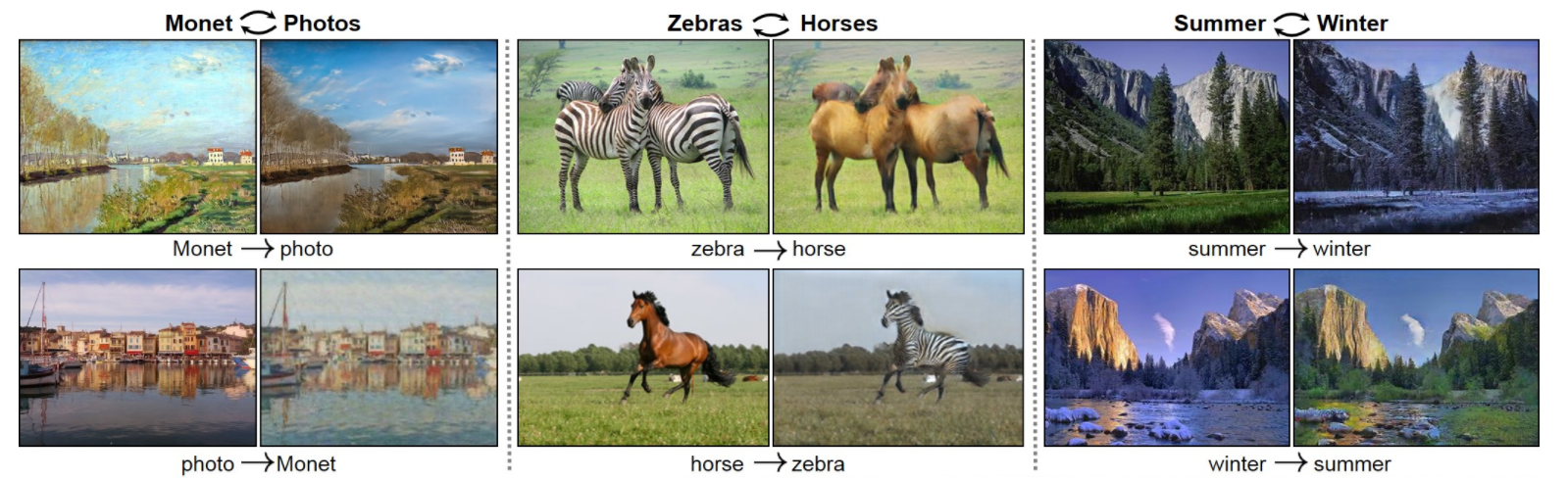}
  \caption{Images generated using CycleGAN~\cite{zhu2017unpaired}}
\end{subfigure}\\ 
\vspace{8pt}
\begin{subfigure}{\textwidth}
  \centering
  \includegraphics[scale=1.12]{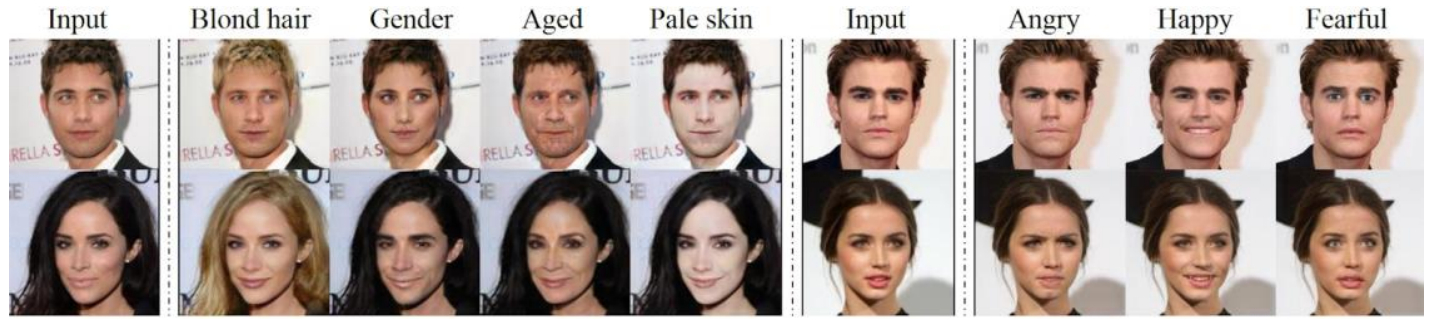}
  \vspace{5pt}
  \caption{Images generated using StarGAN~\cite{choi2018stargan}}
\end{subfigure}
\vspace{14pt}
\caption{Examples of images that have been generated using GANs.}
\label{fig:cyclegan-stargan}
\end{figure*}

GANs have been applied to many image processing tasks such as image synthesis, super-resolution and image completion.
Inspired by the results of these image processing tasks, GANs have brought in novel attack avenues such as computer generated (CG) faces~\cite{karras2017progressive}, augmenting faces with CG facial attributes~\cite{choi2018stargan}, and seamless transfer of texture between images~\cite{zhu2017unpaired}, to name a few.
Two of the most common applications of GANs include texture or style transfer between images and face manipulations.  
An example of GAN generated texture translation between images, such as  horses-to-zebras and summer-to-winter, is shown in Fig.~\ref{fig:cyclegan-stargan}(a).  These techniques manipulates the entire image to change the visual appearance of the scene~\cite{zhu2017unpaired}. 
There has been also tremendous progress in facial manipulations - in particular, automatic generation of facial attributes and expressions.
Fig.~\ref{fig:cyclegan-stargan}(b) shows one such recent example where various facial attributes such as hair, gender, age and skin color, and expressions such as anger, happiness and fear are generated on faces of celebrities~\cite{choi2018stargan}.

While the visual results are promising, the GAN based techniques alter the statistics of pixels in the images that they generate.  
Hence, methods that look for deviations from natural image statistics could be effective in detecting GAN generated fake images.
These methods have been well studied in the field of steganalysis which aims to detect the presence of hidden data in digital images.
One such method is based on analyzing co-occurrences of pixels by computing a co-occurrence matrix.
Traditionally, this method uses hand crafted features computed on the co-occurrence matrix and a machine learning classifier such as support vector machines determines if a message is hidden in the image~\cite{sullivan2005steganalysis,sullivan2006steganalysis}.
Other techniques involve calculating image residuals or passing the image through different filters before computing the co-occurrence matrix~\cite{pevny2010steganalysis,fridrich2012rich, cozzolino2017recasting}.

Inspired by steganalysis and natural image statistics, we propose a novel method to identify GAN generated images using a combination of pixel co-occurrence matrices and deep learning.
Here we pass the co-occurrence matrices directly through a deep learning framework and allow the network to learn important features of the co-occurrence matrices.  
We also avoid computation of residuals or passing an image through various filters, but rather compute the co-occurrence matrices on the image pixels itself.
Experimental results on two diverse and challenging datasets generated using GAN based methods show that our approach is promising and generalizes well when the GAN model is not known during training.

\section{Related Work}
\label{sec:rel-work}

Since the seminal work on GANs~\cite{goodfellow2014generative}, there have been several hundreds of papers on using GANs to generate images. 
These works focus on generating images of high perceptual quality~\cite{mirza2014conditional, radford2015unsupervised,salimans2016improved,isola2017image,arjovsky2017wasserstein,gulrajani2017improved, karras2017progressive}, image-to-image translations~\cite{isola2017image,yi2017dualgan,zhu2017unpaired}, domain transfer~\cite{taigman2016unsupervised,kim2017learning}, super-resolution~\cite{ledig2017photo}, image synthesis and completion~\cite{li2017generative,iizuka2017globally,wang2018high}, and generation of facial attributes and expressions~\cite{liu2016coupled, perarnau2016invertible,kim2017learning,choi2018stargan}.
Several methods have been proposed in the area of image forensics over the past years~\cite{forgery_1,forgery_2,forgery_3,forgery_4,forgery_5}. 
Recent approaches have focused on applying deep learning based methods to detect tampered images~\cite{bayar2016deep, bayar2017design,rao2016deep,bunk2017detection,bappy2017exploiting,cozzolino2017recasting,zhou2018learning}

The detection of GAN images is a new area in image forensics and there are very few papers in this area~\cite{marra2018detection,valle2018tequilagan,li2018detection,mccloskey2018detecting,li2018can,jain2018on,tariq2018detecting,marra2018gans,mo2018fake,do2018forensics}. 
Related fields also include detection of computer generated (CG) images~\cite{dirik2007new,wu2011identifying,mader2017identifying,rahmouni2017distinguishing}.
The most relevant work is a recent paper~\cite{marra2018detection} on detecting GAN based image-to-image translation generated using cycleGAN~\cite{zhu2017unpaired}. 
Here the authors compare various existing methods to identify cycleGAN images from normal ones.
The top results they obtained using a combination of residual features~\cite{cozzolino2014image,cozzolino2017recasting} and deep learning~\cite{chollet2017xception}.
Similar to~\cite{marra2018detection}, the authors in~\cite{li2018detection} compute the residuals of high pass filtered images and then extract co-occurrence matrices on these residuals, which are then concatenated to form a feature vector that can distinguish real from fake GAN images.
In contrast to these approaches, our approach does not need any image residuals to be computed. 
Rather, our method directly computes co-occurrence matrices on the three color channels which are then passed through a deep convolutional neural network (CNN) to learn a model that can detect fake GAN generated images.

\section{Methodology}

\begin{figure*}[t]
\centering
\captionsetup{justification=centering}
\includegraphics[scale=0.5]{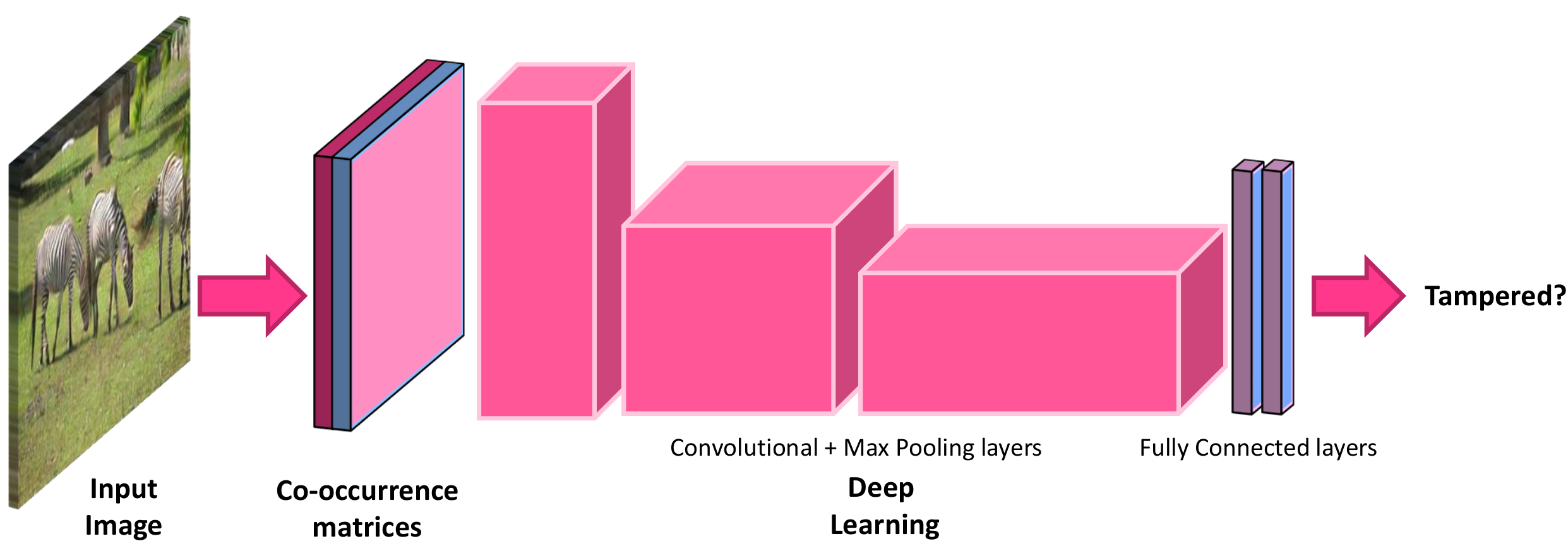}
\vspace{14pt}
\caption{An end-to-end framework to detect GAN generated images} 
\label{fig:ovw}
\end{figure*}

To detect GAN images, we compute co-occurrence matrices on the RGB channels of an image. 
Co-occurrence matrices have been previously used in steganalysis to identify images that have data hidden in them~\cite{sullivan2005steganalysis,sullivan2006steganalysis,pevny2010steganalysis,fridrich2012rich} and in image forensics to detect or localize tampered images~\cite{cozzolino2014image,cozzolino2017recasting}. 
In prior works, co-occurrence matrices are usually computed on image residuals by passing the image through many filters and then obtaining the difference. 
Sometimes features or statistics are also computed on the matrices and then a classifier is trained on these features to classify data hidden or tampered images~\cite{sullivan2005steganalysis}.

However, in this paper, we compute co-occurrence matrices directly on the image pixels on each of the red, green and blue channels and pass them through a convolutional neural network, thereby allowing the network to learn important features from the co-occurrence matrices. An overview of our approach is shown in Fig.~\ref{fig:ovw}. 
Specifically, the first step is to compute the co-occurrence matrices on the RGB channels to obtain a 3x256x256 tensor. 
This tensor is then passed through a multi-layer deep convolutional neural network: conv layer with 32 3x3 convs + ReLu layer + conv layer with 32 5x5 convs + max pooling layer +  conv layer with 64 3x3 convs + ReLu layer + conv layer with 64 5x5 convs + max pooling layer + conv layer with 128 3x3 convs + ReLu layer + conv layer with 128 5x5 convs + max pooling layer + 256 dense layer + 256 dense layer + sigmoid layer. A variant of adaptive stochastic gradient descent is used as the optimizer.

\section{Experiments}

We evaluate our method on two diverse and challenging datasets which contain GAN generated images such as image-to-image translation, style transfer, facial attributes and expressions.

\subsection{Datasets}

\noindent \textbf{CycleGAN dataset:} This dataset contains unpaired image-to-image translations of various objects and scenes such as horses-to-zebras, summer-to-winter, images to paintings (Monet, Van Gough), style transfer such as labels to facades, and others that were generated using a cycle-consistent GAN framework~\cite{zhu2017unpaired}. 
We followed the instructions provided by the authors as detailed in \emph{https://github.com/junyanz/pytorch-CycleGAN-and-pix2pix}, and obtained 36,302 images (18,151 GAN and 18,151 non-GAN images).
The distribution of GAN images are as follows: apple2orange (2014), horse2zebra (2401), summer2winter (2193), cityscapes (2975), facades (400), map2sat
(1096), Ukiyoe (1500), Van Gogh (1500), Cezanne (1500), Monet (2752). 
We obtained this distribution from the authors of~\cite{marra2018detection}\footnote{We thank the authors for providing the dataset distribution} and we compare our approach with their results in the Experiments section.

\noindent \textbf{StarGAN dataset:} This dataset consists of 19,990 images of which 1,999 were from faces taken from the celebA dataset~\cite{liu2015deep} of celebrity faces and the remaining 17,991 images were GAN generated images with 5 varying facial attributes such as black hair, blond hair, brown hair, gender change, aged and 4 combinations of the same.  
We followed the instructions in \emph{https://github.com/yunjey/stargan} to generate the images.

\subsection{Evaluation}
\begin{figure*}[t]
  \centering
  \begin{subfigure}{.24\textwidth}
  \centering
  \includegraphics[scale=0.28]{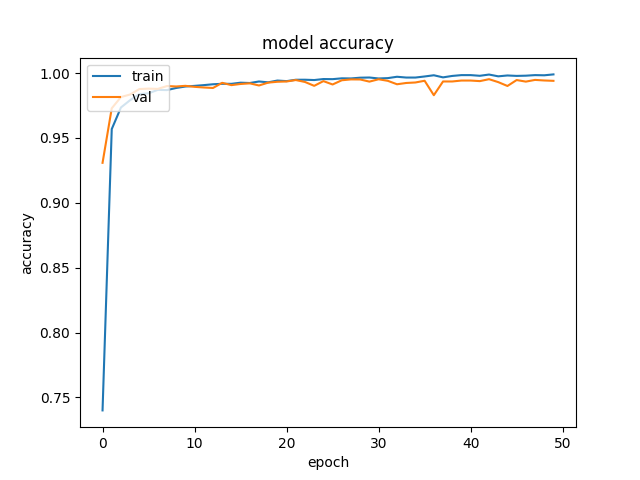}
  \caption{}
\end{subfigure}%
\begin{subfigure}{.24\textwidth}
  \centering
  \includegraphics[scale=0.28]{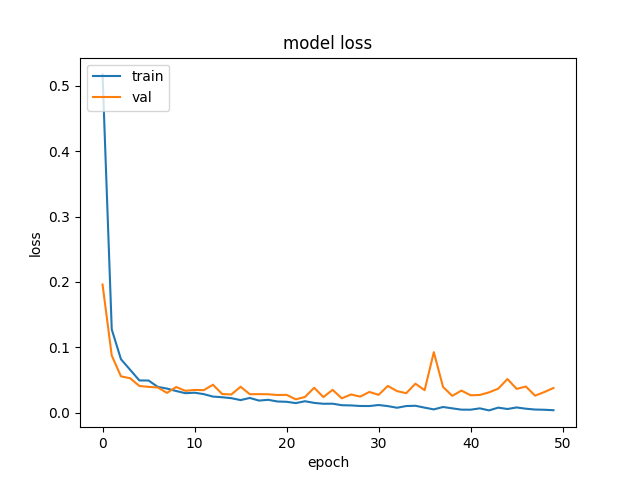}
  \caption{}
\end{subfigure}
\begin{subfigure}{.24\textwidth}
  \centering
  \includegraphics[scale=0.28]{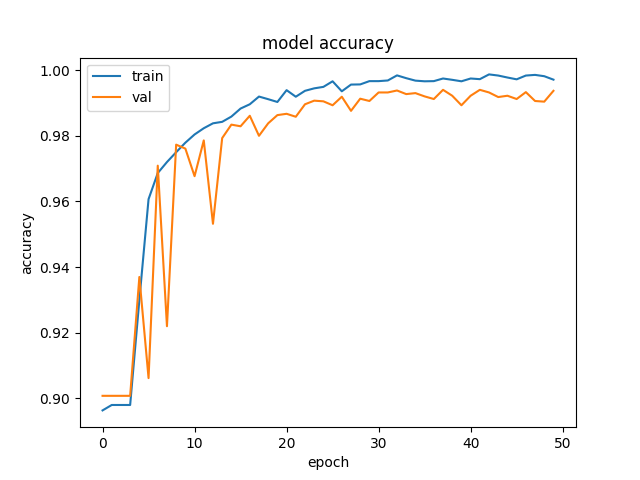}
  \caption{}
\end{subfigure}%
\begin{subfigure}{.24\textwidth}
  \centering
  \includegraphics[scale=0.28]{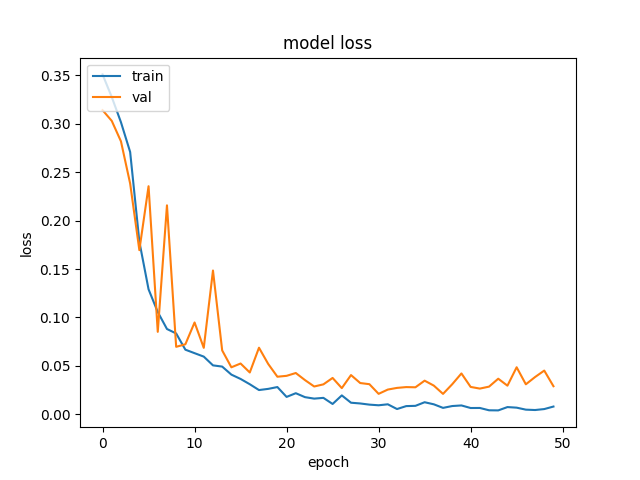}
  \caption{}
\end{subfigure}
\vspace{20pt}
\caption{Model accuracy and loss on CycleGAN dataset~\cite{zhu2017unpaired} (a,b) and StarGAN dataset~\cite{choi2018stargan} (c,d)}
\label{fig:acc-loss}
\end{figure*}

We first evaluate our approach on two datasets separately and then perform cross evaluation on the two datasets (one dataset as training and other as testing) to see the generalizability of our approach.
For both datasets, 50\% of the data is used for training, 25\% for validation and 25\% for testing. 
We train the network for 50 epochs with a batch size of 40 and use a variant of adaptive stochastic gradient as optimizer. 
Fig.~\ref{fig:acc-loss}(a,b) and Fig.~\ref{fig:acc-loss}(c,d) show the model accuracy and loss on cycleGAN dataset and StarGAN dataset, respectively. We obtained a high training and validation accuracy of 99.90\% and 99.40\% on cycleGAN dataset, and  99.43\% and 99.39\% on StarGAN dataset respectively. 
We then evaluated the model on the held-out test sets and obtained a testing accuracy of 99.71\% on the cycleGAN dataset and 99.37\% on the StarGAN dataset.

\subsection{Generalizability}

Next, we evaluate the generalizability of our approach by training on one dataset and testing on the other.
First we train on all the images in the cycleGAN dataset (35,302 images) and test the model on all images of the StarGAN dataset (19,990) images, and then we reverse the experiment where we train on StarGAN and test on cycleGAN.
We train the network till 50 epochs and report on the model that gave the highest accuracy.
As shown in Tab.~\ref{tab:gen}, our method still maintains a high accuracy even across diverse datasets.
The model trained on cycleGAN dataset has a higher accuracy of 99.45\% in comparison with the model trained on StarGAN dataset which got 93.42\%. 
The lower accuracy for the model trained on StarGAN dataset could be because of the non-uniform distribution of class samples in the StarGAN dataset, and due to the diverse image sources/categories in the cycleGAN dataset.

\begin{table}
\centering
\vspace{6pt}
\caption{Experiment on Generalizability}
\vspace{2pt}
\begin{tabular}{|c|c|c|}
\hline
Training dataset & Testing dataset & Accuracy \\
\hline
cycleGAN & StarGAN & 99.49\\
\hline
StarGAN & cycleGAN & 93.42\\
\hline
\end{tabular}
\label{tab:gen}
\end{table}

\begin{table*}
\centering
\vspace{15pt}
\caption{Comparison with State-of-the-art}
\vspace{2pt}
\begin{tabular}{|c|c|c|c|c|c|c|c|c|c|c|c|}
\hline
Method & ap2or & ho2zeb & wint2sum & citysc. & facades & map2sat & Ukiyoe & Van Gogh & Cezanne & Monet & Average \\
\hline
Steganalysis feat. & 98.93 & 98.44 & 66.23 & 100.00 & 97.38 & 88.09 & 97.93 & 99.73 & 99.83 & 98.52 & 94.40\\
\hline
Cozzalino2017 & 99.90 & 99.98 & 61.22 & 99.92 & 97.25 & 99.59 & 100.00 & 99.93 & 100 & 99.16 & 95.07\\
\hline
XceptionNet & 95.91 & 99.16 & 76.74 & 100.00 & 98.56 & 76.79 & 100.00 & 99.93 & 100.00 & 95.10 & 94.49\\
\hline
Proposed & 99.78 & 99.75 & 99.72 & 92.00 & 80.63 & 97.51 & 99.63 & 100.00 & 99.63 & 99.16 & \textbf{97.84}\\
\hline
\end{tabular}
\label{tab:comp1}
\end{table*}

\begin{table*}[!h]
\centering
\vspace{6pt}
\caption{Effect of JPEG compression}
\vspace{2pt}
\begin{tabular}{|c|c|c|}
\hline
JPEG Quality Factor & Trained on original images & Trained on JPEG compressed images \\
\hline
95 & 74.5 & 93.78\\
\hline
85 & 69.46 & 91.61\\
\hline
75 & 64.46 & 87.31\\
\hline
\end{tabular}
\label{tab:jpeg}
\end{table*}
\subsection{Comparison with State-of-the-art}
\label{sec:comparison}
We compare our approach with the results presented in~\cite{marra2018detection}. 
Here the authors conduct a study on the detection of images manipulated by GAN-based image-to-image translation on the cycleGAN dataset.
For evaluation, they adopt a leave-one-manipulation-out strategy on the categories in the cycleGAN dataset, where at each iteration images belonging to one category are set aside for validation and the images from other categories are used for training. 
The methods evaluated are based on steganalysis, generic image manipulations, detection of computer graphics, a GAN discriminator used in the cycleGAN paper, and generic deep learning architecture pretrained on ImageNet~\cite{imagenet_cvpr09}, but fine tuned to the cycleGAN dataset. 
Among these the top preforming ones were from steganalysis~\cite{fridrich2012rich,cozzolino2014image} based on extracting features from high-pass residual images, a deep neural network designed to extract residual features~\cite{cozzolino2017recasting} (denoted by Cozzolino2017) and XceptionNet~\cite{chollet2017xception} deep neural network trained on ImageNet but fine-tuned to this dataset. 
We report the results as mentioned in the paper only from these three methods and compare with our approach. 

\noindent \textbf{Results on original images:} Here we consider the images generated from the cycleGAN dataset as it is and do not perform any postprocessing on the images.
Tab.~\ref{tab:comp1} summarizes the results of our proposed approach along with with the three top performing approaches in~\cite{marra2018detection}.
On average our method outperformed other methods and was able to achieve an accuracy of \textbf{97.84}.
On most categories, our approach was better than or on-par with the other top methods. 
The only categories where our method performed poorly were `cityscapes' and `facades'. 
These could be because the original images in these categories were JPEG compressed which could have affected the classification accuracy. 
In the next section, we study the effect of compression on our method.

\noindent \textbf{Effect of JPEG compression:}
In~\cite{marra2018detection}, the authors investigated the sensitivity of their methods on compression.  Using a compression method similar to Twitter, they trained their detection methods on original uncompressed images and tested on JPEG compressed images.
The objective of this study was to test the robustness of the detection techniques when images are posted in social networks such as Twitter.
In the second scenario, they train and test on the JPEG compressed images.
We performed both of these experiments on the cycleGAN dataset. 
Since we are not aware of the exact JPEG quantization tables used in Twitter, our approach was similar but we tested on three different JPEG quality factors (QF): 95, 85 and 75.
We used 50\% of the data for training, 25\% for validation and 25\% for testing. 
The results are reported on the 25\% testing data of the cycleGAN dataset (9,076 images).
As shown in Tab.~\ref{tab:jpeg}, the accuracy progressively drops as the QF decreases from 95-75, when trained on the original images.
But when the JPEG compressed images are used for training, the accuracy shows a substantial increase.
Even at a QF of 75, the accuracy is still 87.31\%.
This is also consistent with the results reported in~\cite{marra2018detection}, where they report close to a 10\% drop in accuracy on Twitter-like compressed images.


\section{Conclusions}

In this paper, we proposed a novel method to detect GAN generated fake images using a combination of pixel co-occurrence matrices and deep learning.
Co-occurrence matrices are computed on the color channels of an image and then trained using a deep convolutional neural network to distinguish GAN generated fake images from real ones.
Experiments on two diverse GAN datasets show that our approach is both effective and generalizable. 
In future, we will consider localizing the manipulated pixels in GAN generated fake images.

\section{Acknowledgments} 

This research was developed with funding from the Defense Advanced Research Projects Agency (DARPA).
The views, opinions and/or findings expressed are those of the author and should not be interpreted as representing the official views or policies of the Department of Defense or the U.S. Government. 
The paper is approved for public release, distribution unlimited.

{\small
	\bibliographystyle{spiejour}
	\bibliography{gan-detection}

\begin{thebibliography}{10}

\bibitem{zhu2017unpaired}
J.-Y. Zhu, T.~Park, P.~Isola, {\em et~al.}, ``Unpaired image-to-image
  translation using cycle-consistent adversarial networks,'' in {\em IEEE
  International Conference on Computer Vision},   (2017).

\bibitem{choi2018stargan}
Y.~Choi, M.~Choi, M.~Kim, {\em et~al.}, ``Stargan: Unified generative
  adversarial networks for multi-domain image-to-image translation,'' in {\em
  Proceedings of the IEEE Conference on Computer Vision and Pattern
  Recognition},  8789--8797  (2018).

\bibitem{goodfellow2014generative}
I.~Goodfellow, J.~Pouget-Abadie, M.~Mirza, {\em et~al.}, ``Generative
  adversarial nets,'' in {\em Advances in neural information processing
  systems},  2672--2680  (2014).

\bibitem{karras2017progressive}
T.~Karras, T.~Aila, S.~Laine, {\em et~al.}, ``Progressive growing of gans for
  improved quality, stability, and variation,'' {\em arXiv preprint
  arXiv:1710.10196}   (2017).

\bibitem{sullivan2005steganalysis}
K.~Sullivan, U.~Madhow, S.~Chandrasekaran, {\em et~al.}, ``Steganalysis of
  spread spectrum data hiding exploiting cover memory,'' in {\em Security,
  Steganography, and Watermarking of Multimedia Contents VII},   {\bf 5681},
  38--47, International Society for Optics and Photonics  (2005).

\bibitem{sullivan2006steganalysis}
K.~Sullivan, U.~Madhow, S.~Chandrasekaran, {\em et~al.}, ``Steganalysis for
  markov cover data with applications to images,'' {\em IEEE Transactions on
  Information Forensics and Security} {\bf 1}(2), 275--287  (2006).

\bibitem{pevny2010steganalysis}
T.~Pevn{\`y}, P.~Bas, and J.~Fridrich, ``Steganalysis by subtractive pixel
  adjacency matrix,'' {\em information Forensics and Security, IEEE
  Transactions on} {\bf 5}(2), 215--224  (2010).

\bibitem{fridrich2012rich}
J.~Fridrich and J.~Kodovsky, ``Rich models for steganalysis of digital
  images,'' {\em IEEE Transactions on Information Forensics and Security} {\bf
  7}(3), 868--882  (2012).

\bibitem{cozzolino2017recasting}
D.~Cozzolino, G.~Poggi, and L.~Verdoliva, ``Recasting residual-based local
  descriptors as convolutional neural networks: an application to image forgery
  detection,'' in {\em Proceedings of the 5th ACM Workshop on Information
  Hiding and Multimedia Security},  159--164, ACM  (2017).

\bibitem{mirza2014conditional}
M.~Mirza and S.~Osindero, ``Conditional generative adversarial nets,'' {\em
  arXiv preprint arXiv:1411.1784}   (2014).

\bibitem{radford2015unsupervised}
A.~Radford, L.~Metz, and S.~Chintala, ``Unsupervised representation learning
  with deep convolutional generative adversarial networks,'' {\em arXiv
  preprint arXiv:1511.06434}   (2015).

\bibitem{salimans2016improved}
T.~Salimans, I.~Goodfellow, W.~Zaremba, {\em et~al.}, ``Improved techniques for
  training gans,'' in {\em Advances in Neural Information Processing Systems},
  2234--2242  (2016).

\bibitem{isola2017image}
P.~Isola, J.-Y. Zhu, T.~Zhou, {\em et~al.}, ``Image-to-image translation with
  conditional adversarial networks,'' in {\em Proceedings of the IEEE
  Conference on Computer Vision and Pattern Recognition},  1125--1134  (2017).

\bibitem{arjovsky2017wasserstein}
M.~Arjovsky, S.~Chintala, and L.~Bottou, ``Wasserstein gan,'' {\em arXiv
  preprint arXiv:1701.07875}   (2017).

\bibitem{gulrajani2017improved}
I.~Gulrajani, F.~Ahmed, M.~Arjovsky, {\em et~al.}, ``Improved training of
  wasserstein gans,'' in {\em Advances in Neural Information Processing
  Systems},  5767--5777  (2017).

\bibitem{yi2017dualgan}
Z.~Yi, H.~Zhang, P.~Tan, {\em et~al.}, ``Dualgan: Unsupervised dual learning
  for image-to-image translation,'' in {\em 2017 IEEE International Conference
  on Computer Vision (ICCV)},  2868--2876, IEEE  (2017).

\bibitem{taigman2016unsupervised}
Y.~Taigman, A.~Polyak, and L.~Wolf, ``Unsupervised cross-domain image
  generation,'' {\em arXiv preprint arXiv:1611.02200}   (2016).

\bibitem{kim2017learning}
T.~Kim, M.~Cha, H.~Kim, {\em et~al.}, ``Learning to discover cross-domain
  relations with generative adversarial networks,'' in {\em International
  Conference on Machine Learning},  1857--1865  (2017).

\bibitem{ledig2017photo}
C.~Ledig, L.~Theis, F.~Huszar, {\em et~al.}, ``Photo-realistic single image
  super-resolution using a generative adversarial network,'' in {\em
  Proceedings of the IEEE Conference on Computer Vision and Pattern
  Recognition},  4681--4690  (2017).

\bibitem{li2017generative}
Y.~Li, S.~Liu, J.~Yang, {\em et~al.}, ``Generative face completion,'' in {\em
  The IEEE Conference on Computer Vision and Pattern Recognition (CVPR)},
  {\bf 1}(2), 3  (2017).

\bibitem{iizuka2017globally}
S.~Iizuka, E.~Simo-Serra, and H.~Ishikawa, ``Globally and locally consistent
  image completion,'' {\em ACM Transactions on Graphics (TOG)} {\bf 36}(4), 107
   (2017).

\bibitem{wang2018high}
T.-C. Wang, M.-Y. Liu, J.-Y. Zhu, {\em et~al.}, ``High-resolution image
  synthesis and semantic manipulation with conditional gans,'' in {\em
  Proceedings of the IEEE Conference on Computer Vision and Pattern
  Recognition},  8798--8807  (2018).

\bibitem{liu2016coupled}
M.-Y. Liu and O.~Tuzel, ``Coupled generative adversarial networks,'' in {\em
  Advances in neural information processing systems},  469--477  (2016).

\bibitem{perarnau2016invertible}
G.~Perarnau, J.~van~de Weijer, B.~Raducanu, {\em et~al.}, ``Invertible
  conditional gans for image editing,'' {\em arXiv preprint arXiv:1611.06355}
  (2016).

\bibitem{forgery_1}
H.~Farid, ``Image forgery detection,'' {\em IEEE Signal processing magazine}
  {\bf 26}(2), 16--25  (2009).

\bibitem{forgery_2}
B.~Mahdian and S.~Saic, ``A bibliography on blind methods for identifying image
  forgery,'' {\em Signal Processing: Image Communication} {\bf 25}(6), 389--399
   (2010).

\bibitem{forgery_3}
G.~K. Birajdar and V.~H. Mankar, ``Digital image forgery detection using
  passive techniques: A survey,'' {\em Digital Investigation} {\bf 10}(3),
  226--245  (2013).

\bibitem{forgery_4}
X.~Lin, J.-H. Li, S.-L. Wang, {\em et~al.}, ``Recent advances in passive
  digital image security forensics: A brief review,'' {\em Engineering}
  (2018).

\bibitem{forgery_5}
S.~Walia and K.~Kumar, ``Digital image forgery detection: a systematic
  scrutiny,'' {\em Australian Journal of Forensic Sciences} , 1--39  (2018).

\bibitem{bayar2016deep}
B.~Bayar and M.~C. Stamm, ``A deep learning approach to universal image
  manipulation detection using a new convolutional layer,'' in {\em Proceedings
  of the 4th ACM Workshop on Information Hiding and Multimedia Security},
  5--10  (2016).

\bibitem{bayar2017design}
B.~Bayar and M.~C. Stamm, ``Design principles of convolutional neural networks
  for multimedia forensics,'' in {\em The 2017 IS\&T International Symposium on
  Electronic Imaging: Media Watermarking, Security, and Forensics},  IS\&T
  Electronic Imaging  (2017).

\bibitem{rao2016deep}
Y.~Rao and J.~Ni, ``A deep learning approach to detection of splicing and
  copy-move forgeries in images,'' in {\em Information Forensics and Security
  (WIFS), 2016 IEEE International Workshop on},  1--6, IEEE  (2016).

\bibitem{bunk2017detection}
J.~Bunk, J.~H. Bappy, T.~M. Mohammed, {\em et~al.}, ``Detection and
  localization of image forgeries using resampling features and deep
  learning,'' in {\em Computer Vision and Pattern Recognition Workshops
  (CVPRW), 2017 IEEE Conference on},  1881--1889, IEEE  (2017).

\bibitem{bappy2017exploiting}
J.~H. Bappy, A.~K. Roy-Chowdhury, J.~Bunk, {\em et~al.}, ``Exploiting spatial
  structure for localizing manipulated image regions,'' in {\em Proceedings of
  the IEEE International Conference on Computer Vision},   (2017).

\bibitem{zhou2018learning}
P.~Zhou, X.~Han, V.~I. Morariu, {\em et~al.}, ``Learning rich features for
  image manipulation detection,'' {\em arXiv preprint arXiv:1805.04953}
  (2018).

\bibitem{marra2018detection}
F.~Marra, D.~Gragnaniello, D.~Cozzolino, {\em et~al.}, ``Detection of
  gan-generated fake images over social networks,'' in {\em 2018 IEEE
  Conference on Multimedia Information Processing and Retrieval (MIPR)},
  384--389, IEEE  (2018).

\bibitem{valle2018tequilagan}
R.~Valle, U.~CNMAT, W.~Cai, {\em et~al.}, ``Tequilagan: How to easily identify
  gan samples,'' {\em arXiv preprint arXiv:1807.04919}   (2018).

\bibitem{li2018detection}
H.~Li, B.~Li, S.~Tan, {\em et~al.}, ``Detection of deep network generated
  images using disparities in color components,'' {\em arXiv preprint
  arXiv:1808.07276}   (2018).

\bibitem{mccloskey2018detecting}
S.~McCloskey and M.~Albright, ``Detecting gan-generated imagery using color
  cues,'' {\em arXiv preprint arXiv:1812.08247}   (2018).

\bibitem{li2018can}
H.~Li, H.~Chen, B.~Li, {\em et~al.}, ``Can forensic detectors identify gan
  generated images?,'' in {\em APSIPA Annual Summit and Conference 2018},
  (2018).

\bibitem{jain2018on}
A.~Jain, R.~Singh, and M.~Vatsa, ``On detecting gans and retouching based
  synthetic alterations,'' in {\em 9th International Conference on Biometrics:
  Theory, Applications and Systems},   (2018).

\bibitem{tariq2018detecting}
S.~Tariq, S.~Lee, H.~Kim, {\em et~al.}, ``Detecting both machine and human
  created fake face images in the wild,'' in {\em Proceedings of the 2nd
  International Workshop on Multimedia Privacy and Security},  81--87, ACM
  (2018).

\bibitem{marra2018gans}
F.~Marra, D.~Gragnaniello, L.~Verdoliva, {\em et~al.}, ``Do gans leave
  artificial fingerprints?,'' {\em arXiv preprint arXiv:1812.11842}   (2018).

\bibitem{mo2018fake}
H.~Mo, B.~Chen, and W.~Luo, ``Fake faces identification via convolutional
  neural network,'' in {\em Proceedings of the 6th ACM Workshop on Information
  Hiding and Multimedia Security},  43--47, ACM  (2018).

\bibitem{do2018forensics}
N.-T. Do, I.-S. Na, and S.-H. Kim, ``Forensics face detection from gans using
  convolutional neural network,'' in {\em ISITC'2018},   (2018).

\bibitem{dirik2007new}
A.~E. Dirik, S.~Bayram, H.~T. Sencar, {\em et~al.}, ``New features to identify
  computer generated images,'' in {\em Image Processing, 2007. ICIP 2007. IEEE
  International Conference on},   {\bf 4}, IV--433, IEEE  (2007).

\bibitem{wu2011identifying}
R.~Wu, X.~Li, and B.~Yang, ``Identifying computer generated graphics via
  histogram features,'' in {\em Image Processing (ICIP), 2011 18th IEEE
  International Conference on},  1933--1936, IEEE  (2011).

\bibitem{mader2017identifying}
B.~Mader, M.~S. Banks, and H.~Farid, ``Identifying computer-generated
  portraits: The importance of training and incentives,'' {\em Perception} {\bf
  46}(9), 1062--1076  (2017).

\bibitem{rahmouni2017distinguishing}
N.~Rahmouni, V.~Nozick, J.~Yamagishi, {\em et~al.}, ``Distinguishing computer
  graphics from natural images using convolution neural networks,'' in {\em
  Information Forensics and Security (WIFS), 2017 IEEE Workshop on},  1--6,
  IEEE  (2017).

\bibitem{cozzolino2014image}
D.~Cozzolino, D.~Gragnaniello, and L.~Verdoliva, ``Image forgery detection
  through residual-based local descriptors and block-matching,'' in {\em Image
  Processing (ICIP), 2014 IEEE International Conference on},  5297--5301, IEEE
  (2014).

\bibitem{chollet2017xception}
F.~Chollet, ``Xception: Deep learning with depthwise separable convolutions,''
  {\em arXiv preprint} , 1610--02357  (2017).

\bibitem{liu2015deep}
Z.~Liu, P.~Luo, X.~Wang, {\em et~al.}, ``Deep learning face attributes in the
  wild,'' in {\em Proceedings of the IEEE International Conference on Computer
  Vision},  3730--3738  (2015).

\bibitem{imagenet_cvpr09}
J.~Deng, W.~Dong, R.~Socher, {\em et~al.}, ``{ImageNet: A Large-Scale
  Hierarchical Image Database},'' in {\em CVPR09},   (2009).

\end{thebibliography}
}


\begin{biography}

\textbf{Lakshmanan Nataraj} received his B.E degree in Electronics and Communications Engineering from Sri Venkateswara College of Engineering (affiliated to Anna University) in 2007, and the Ph.D. degree in the Electrical and Computer Engineering from the University of California, Santa Barbara in 2015. 
He is currently a Senior Research Staff Member at Mayachitra Inc., Santa Barbara, CA, where he leads research projects on malware detection and media forensics. 
His research interests include multimedia security, malware detection and image forensics.

\textbf{Tajuddin Manhar Mohammed} received his B.Tech (Hons.) degree in Electrical Engineering from Indian Institute of Technology (IIT), Hyderabad, India in 2015 and his M.S. degree in Electrical and Computer Engineering from University of California Santa Barbara (UCSB), Santa Barbara, CA in 2016. After obtaining his Masters degree, he obtained a job as a Research Staff Member for Mayachitra Inc., Santa Barbara, CA. His recent research efforts include developing deep learning and computer vision techniques for image forensics and cyber security.

\textbf{B. S. Manjunath} (S'88–M'91–SM'01–F'05) received the Ph.D. degree in electrical engineering from the University of Southern California, Los Angeles, CA, USA, in 1991. He is currently a Distinguished Professor of Electrical and Computer Engineering at  the University of California at Santa Barbara, where he directs the Center for Multimodal Big Data Science and Healthcare. He has authored or coauthored about 300 peer-reviewed articles and served as an Associate Editor for the IEEE Transactions on Image Processing, the IEEE Transactions on Pattern Analysis and Machine Intelligence, the IEEE Transactions on Multimedia, the IEEE Transactions on Information Forensics and Security, and the IEEE Signal Processing Letters.  His research interests include image processing, machine learning, computer vision, and media forensics. He is a fellow of IEEE and ACM.

\textbf{Shivkumar Chandrasekaran} received his M.Sc. (Hons.)
degree in physics from the Birla Institute of Technology and Science (BITS), Pilani, India, in 1987,
and his Ph.D. degree in Computer Science from Yale
University, New Haven, CT, in 1994.
He was a Visiting Instructor at North Carolina
State University, Raleigh, in the Mathematics Department,
before joining the Electrical and Computer
Engineering Department, University of California,
Santa Barbara, where he is currently a Professor. His
research interests are in Computational Mathematics

\textbf{Arjuna Flenner} received his Ph.D. in Physics at the University of Missouri-Columbia located in Columbia MO in the year 2004. His major emphasis was mathematical Physics. After obtaining his Ph.D., Arjuna Flenner obtained a job as a research physicist for NAVAIR at China Lake CA. He won the 2013 Dr. Delores M. Etter Navy Scientist and Engineer award for his work on Machine Learning. 

\textbf{Jawadul H. Bappy} received the B.S. degree in Electrical and Electronic Engineering from the Bangladesh University of Engineering and Technology, Dhaka in 2012.
He received his Ph.D. in Electrical and Computer Engineering from the University of California, Riverside in 2018.
He is currently working as a scientist at JD.Com in Mountain View, CA.
His main research interest includes media forensics, deep generative models, and advanced machine learning techniques for real-life applications.

.

\textbf{Amit K. Roy-Chowdhury} received the Bachelor’s degree in Electrical Engineering from Jadavpur University, Calcutta, India, the Master’s degree in Systems Science and Automation from the Indian Institute of Science, Bangalore, India, and the Ph.D. degree in Electrical and Computer Engineering from the University of Maryland, College Park. He is a Professor of Electrical and Computer Engineering and a Cooperating Faculty in the Department of Computer Science and Engineering, University of California, Riverside. His broad research interests include computer vision, image processing, and vision-based statistical learning, with applications in cyber-physical, autonomous and intelligent systems. He is a coauthor of two books: Camera Networks: The Acquisition and Analysis of Videos over Wide Areas, and Recognition of Humans and Their Activities Using Video. He is the editor of the book Distributed Video Sensor Networks. He has been on the organizing and program committees of multiple computer vision and image processing conferences and is serving on the editorial boards of multiple journals. He is a Fellow of the IEEE and IAPR.

\end{biography}

\end{document}